\documentclass[10pt, a4paper]{article}

\usepackage{lrec-coling2024} % this is the new style

\usepackage{listings}
\usepackage{natbib}
\usepackage{multibib}
\usepackage{graphicx}
\usepackage{tabularx}
\usepackage{soul}
\usepackage{xcolor}
\usepackage{hyperref}
\usepackage{utfsym}
\usepackage{xstring}
\usepackage{booktabs}
\usepackage{color}
\usepackage{makecell}
\usepackage{multirow}
\usepackage{latexsym}
\usepackage{amssymb}
\usepackage{amsfonts}
\usepackage{amsmath} 
\usepackage{amsthm} 
\newcommand{\ie}{\emph{i.e.}, }
\newcommand{\eg}{\emph{e.g.}, }

\newcommand{\paratitle}[1]{\vspace{1.5ex}\noindent\textbf{#1}}

\title{ChainLM: Empowering Large Language Models with Improved Chain-of-Thought Prompting}

\name{Xiaoxue Cheng\textsuperscript{\rm{1}}, Junyi Li\textsuperscript{\rm{1,3}}, Wayne Xin Zhao\textsuperscript{\rm{1}$^{\ast}$\thanks{*Corresponding author}} {\rm and} Ji-Rong Wen\textsuperscript{\rm{1,2}}} 

\address{\textsuperscript{1}Gaoling School of Artificial Intelligence, Renmin University of China \\
\textsuperscript{2}School of Information, Renmin University of China \\
\textsuperscript{3}DIRO, Universit\'{e} de Montr\'{e}al  \\
         \{chengxiaoxue, lijunyi\}@ruc.edu.cn, batmanfly@gmail.com\\}

\abstract{
 Chain-of-Thought (CoT) prompting can enhance the reasoning capabilities of large language models (LLMs), establishing itself as a primary approach to solving complex reasoning tasks. 
 Existing CoT synthesis approaches usually focus on simpler reasoning tasks and thus result in low-quality and inconsistent CoT prompts.
 In response to this challenge, we present an empirical investigation of CoT prompting and introduce CoTGenius, a novel framework designed for the automatic generation of superior CoT prompts. CoTGenius is developed based on three major evolution strategies, \ie complicate, diversify, and specify—alongside two filtering mechanisms: evolutionary success judgement and correctness verification. We further employ CoTGenius to create an extensive CoT dataset, and subsequently fine-tune the Llama 2-Chat 7B and 13B models on this dataset. We call the resulting model ChainLM. 
 To deal with the cumulative error issue in reasoning steps, we propose a step-level debating method, wherein multiple debaters discuss each reasoning step to arrive at the correct answer.
Extensive experiments demonstrate that our ChainLM models exhibit enhanced proficiency in addressing a spectrum of complex reasoning problems compared to existing models. 
In addition, we conduct an in-depth analysis of the impact of data categories within CoTGenius on the model performance. 
We release our dataset and code at \url{https://github.com/RUCAIBox/ChainLM}.
 \\ \newline \Keywords{Chain-of-Thought, Large Language Models, Debating} 
}

\begin{document}

\maketitleabstract
\section{Introduction}
\label{sec-intro}

Large Language Models (LLMs)~\cite{LLM-survey}  have recently made great progress on natural language understanding and generation, and have also shown great potential as general-purpose task solvers by 
following commands or prompts~\cite{Brown-NeurIPS-2020-Language,Nakano-arxiv-2021-WebGPT,OpenAI-OpenAI-2023-GPT-4}. However, 
it is well known that LLMs often fall short in complex reasoning tasks, \eg mathematical reasoning and symbolic reasoning.
Thus
concerns about improving the reasoning capabilities of LLMs have drawn significant attention from the research community~\cite{WizardMath,Logic-LM}.

\begin{table*}[!h]
\centering
\small
\resizebox{\textwidth}{!}{
\begin{tabular}{l|cccc}
\toprule
% \textbf{Methods} & \textbf{Type} & \textbf{Strategy} & \textbf{used for}&\textbf{Correctness check}\\
Methods  & Generation strategy & Task-agnostic & LLM fine-tuning & Correctness check\\
\midrule
Few-shot CoT~\cite{wei2022chain} & NA & \usym{2717} & \usym{2717} & \usym{2717} \\
% \midrule
Zero-shot CoT~\cite{kojima2022large} & NA & \usym{2713}  & \usym{2717} & \usym{2717} \\
Auto-CoT~\cite{zhang2022automatic} &question clustering & \usym{2717} & \usym{2717} & \usym{2717} \\
CoT Collection~\cite{kim2023cot}& human-crafted \& ICL & \usym{2717}  & \usym{2713} & \usym{2717} \\
% \midrule
CoTGenius~(ours) &  complicate, diversify, specify & \usym{2713}  &\usym{2713} & \usym{2713}\\
\bottomrule
\end{tabular}
}
\caption{A comparison between our CoTGenius and other CoT generation methods.}
\label{tab: cot}
\vspace{-0.2cm}
\end{table*}

Amidst this backdrop, Chain-of-Thought (CoT) prompting has been proposed and emerged as an effective solution for complex reasoning~\cite{wei2022chain}, where LLMs incorporate a series of intermediate reasoning steps before inferring the final output. 
\citet{kojima2022large} simply add a phrase ``\emph{Let's think step by step}'' in prompts, enabling LLMs to conduct zero-shot CoT reasoning without any additional exemplars. 
However, the same benefits may not necessarily extend to relatively smaller LLMs.
Although there have been some attempts towards \emph{fine-tuning} LLMs with multi-step CoT prompting data to stimulate the step-by-step reasoning capacity in smaller models~\cite{Shridhar-2022-Distilling,FuPOSK23,kim2023cot}, these approaches are mainly focused on single reasoning tasks or rely on simple strategies to synthesize CoT prompting data. 
\citet{FuPSCK23} found that existing datasets mostly focus on simple tasks involving fewer reasoning steps (\ie only 2 or 3 steps) or even omit certain intermediate steps, making the reasoning process incomplete and the model struggles in complex reasoning tasks.
In addition, these automatic methods ignore considering the consistency between the reasoning process and the final answer~\cite{0002WSLCNCZ23}, resulting in low-quality and spurious CoT prompting data.
Considering these issues, we aim to improve upon existing CoT prompting data for enhancing the complex reasoning capacities of LLMs.
Different from existing efforts that primarily improve instructions with evolutionary algorithms~\cite{xu2023wizardlm,Guo-2023-arxiv-Connecting}, CoT improvement requires considering both the question and reasoning steps, as well as their consistency.
However, the underlying mechanics and efficacy of CoT prompts for LLMs remain underexplored and we would like to investigate a more sophisticated and fundamental question: \emph{what kind of CoT chains can more effectively elicit the potential capacities for LLMs?}

To delve into this problem, we conduct a series of empirical analysis experiments on GSM8K~\citeplanguageresource{Cobbe-arxiv-2021-Training} to study the impact of CoT from three main aspects, \ie completeness, specificity, and logicality. 
First, we vary the number of reasoning steps to examine the performance of CoT prompts with different levels of reasoning completeness. Second, we iteratively refine the CoT prompts to make the reasoning processes more detailed and specific. Finally, we explore the logicality of CoT prompts by interchanging the order of reasoning processes and the final answer. Based on the empirical results, we find that incorporating additional and more specific reasoning steps contributes to a highly 
complete chain-of-thought, resulting in more accurate answers and better performance in mathematical problems. 
Finally, the logic of reasoning before answering, used in the majority of prior studies, leads to substantial performance improvements compared to the logic of answering then explaining.
Based on these findings, we aim to improve upon existing CoT prompting data and fine-tune LLMs on newly generated high-quality CoT data to enhance complex reasoning abilities of LLMs.

There are three major technical contributions in our work. 
First, we propose CoTGenius, a Chain-of-Thought prompting improvement framework to synthesize more complicated, diverse, and detailed CoT prompts, as shown in Figure~\ref{fig-model}. In this framework, we introduce three evolution strategies to improve CoT prompts, \ie \emph{complicate}, \emph{diversify}, and \emph{specify}. 
Starting from a simple initial CoT prompt (including the question and reasoning steps), the complication and diversity strategies will first upgrade the initial question to a more complex one and create a new diverse question, and then generate reasoning steps for the evolved question. The specificity strategy will rewrite the initial CoT reasoning steps and incorporate more details. 
Since the CoT improvement is completely conducted by LLMs, the resulting CoT prompting data might be erroneous. Thus we design two filtering strategies to filter the failed CoT prompts, \ie \emph{evolutionary success judgement} and \emph{correctness verification}. 
We compared CoTGenius with previous CoT generation work in Table~\ref{tab: cot}.
We repeat the improvement process for four rounds using OpenAI ChatGPT API and finally obtain 44,335 CoT prompts.

Second, to validate the effectiveness of our CoTGenius framework, we fine-tune open-source LLMs (\ie Llama 2-Chat 7B and 13B) with our evolved CoT data, called \emph{ChainLM}. We compare ChainLM to existing popular LLMs on several complex reasoning tasks and our model surpasses many open-source LLMs (\eg Alpaca, Vicuna) with significant improvements. 
Specially, ChainLM is superior to directly fine-tuning LLMs on existing simple CoT datasets in large scale~\cite{kim2023cot}, which contain about 1.88 million CoT rationales.
These datasets are simply built upon existing NLP tasks using prompts without considering the complication and specificity of CoT rationales.

Finally, to deal with the cumulative error issue in reasoning steps leading to inaccurate answers, we propose a CoT reasoning strategy, \emph{step-level debating}. This method employs multiple agents (\ie LLMs) to debate about each intermediate reasoning step for a consensus, which can capitalize on the strengths of different models to improve the accuracy of intermediate steps.
Compared to previous CoT reasoning strategies such as self-consistency and least-to-most, our step-level debating method based on ChainLM exhibits better performance on several reasoning tasks.

\section{Empirical Analysis}
\label{sec-emp}

As CoT prompting gradually becomes an effective means of solving complex reasoning tasks, we try to investigate what kind of CoT chains can more effectively elicit the potential capacities for LLMs from three aspects: inference completeness, prompt specificity, and reasoning logicality. All experiments are conducted by ChatGPT in GSM8K dataset.

\subsection{Inference Completeness of CoT}

As discovered in prior work~\cite{FuPSCK23, WizardMath}, the current CoT prompts usually consist of simpler and incomplete reasoning steps (\ie only 2 or 3 steps), which poses challenges for LLMs when tackling complex reasoning tasks.
Since the step-by-step reasoning plays a pivotal role in guiding LLMs to arrive at answers, we aim to explore the influence of the completeness of reasoning steps (\ie the number of reasoning steps). Specially, we instruct ChatGPT with 2, 3 or 5 reasoning steps to solve mathematical problems in GSM8K. 

As shown in Table~\ref{tab: steps}, the results provide clear evidence of a positive correlation between the number of reasoning steps and model accuracy.
Intuitively, as more intermediate reasoning steps are incorporated, LLMs are capable of inferring the final answer with greater ease and accuracy.
This observation aligns well with our expectations and underscores the significance of completing step-wise reasoning in the CoT framework.

\subsection{Prompt Specificity of CoT}

Besides the number of reasoning steps, the level of specificity and details for each step also play pivot roles in CoT reasoning. If the CoT reasoning process misses some important details about the question, LLMs may conduct superficial reasoning, leading to inaccurate answers. To explore the efficacy of the specificity of CoT prompts, we employ ChatGPT to iteratively refine existing CoT reasoning steps to be more specific and detailed. 
We perform two refinement iterations and the results are presented in Table~\ref{tab: iteration}.

As we can see from the table, after one iteration of specificity refinement, the model could achieve satisfactory performance and additional iterations bring minimal accuracy benefits. The reasons behind might be that appropriately adding details to the reasoning process is able to assist LLMs in inferring answers more accurately, while more specificity iterations introduces little useful information to the model's reasoning. These findings underscore the significance of increasing the specificity of CoT prompts, ultimately contributing to the enhancement of CoT reasoning and problem-solving proficiency.

\begin{table}[t]
\centering
\small
\setlength\tabcolsep{12pt}
\begin{tabular}{l|ccc}
\toprule
\textbf{Steps} & \textbf{2} & \textbf{3} & \textbf{5} \\
\midrule
\textbf{Accuracy} & 45.81 & 53.55 & 63.23 \\
\bottomrule
\end{tabular}
\caption{Inference completeness: Accuracy (\%) of different numbers of reasoning steps on GSM8K.}
\label{tab: steps}
% \vspace{-0.2cm}
\end{table}

\begin{table}[t]
\centering
\small
\setlength\tabcolsep{12pt}
\begin{tabular}{l|ccc}
\toprule
\textbf{Iterations} & \textbf{0} & \textbf{1} & \textbf{2} \\
\midrule
\textbf{Accuracy} & 76.49 & 79.15 & 79.37 \\
\bottomrule
\end{tabular}
\caption{Prompt specificity: Accuracy (\%) of varying refinement iterations on GSM8K.}
\label{tab: iteration}
\vspace{-0.2cm}
\end{table}

\begin{table}[t]
\centering
\small
\setlength\tabcolsep{14pt}
\begin{tabular}{l|cc}
\toprule
\textbf{Answer Position} & \textbf{front} & \textbf{behind} \\
\midrule
\textbf{Accuracy} & 68.69 & 76.80 \\
\bottomrule
\end{tabular}
\caption{Reasoning logicality: Accuracy (\%) of different answer positions on GSM8K.}
\label{tab: position}
\vspace{-0.2cm}
\end{table}

\subsection{Reasoning Logicality of CoT}

When employing the CoT prompting to guide LLMs, most studies adopt the logic of first conducting step-by-step reasoning and then deriving the final answer based on the previous reasoning steps. However, in our empirical experiments, we explore another logic of first providing the answer at the outset, followed by detailed step-by-step explanations. This way can be formulated as an outcome explanation problem in explainable AI, which is called \emph{rationalization}~\cite{Rationalization-survey}.

In our experiments, we aim to investigate two logical patterns of CoT, \ie inferring the answer at the front or behind of the reasoning steps, and the results are present in Table~\ref{tab: position}. 
As can be seen from the results, the commonly used CoT logic of predicting the answer after rigorous reasoning, a structure more aligned with human thinking patterns, yields higher accuracy.
This observation underscores the significance of mirroring human cognitive processes in the CoT framework to enhance model performance in complex reasoning tasks by using the logic of reasoning-then-answering.

\hspace*{\fill}

In the above empirical experiments, we have observed that there are three factors that affect CoT reasoning for LLMs. \textbf{First}, increasing the number of reasoning steps plays a pivotal role in enhancing the completeness of CoT prompts and leading to accurate solutions. \textbf{Second}, improving the level of specificity and details for CoT prompts can assist LLMs in performing rigorous step-by-step reasoning and making accurate predictions. \textbf{Third}, the logic of reasoning-then-answering is an intuitive and more effective thinking pattern for CoT reasoning.
In the next section, we will improve existing CoT prompts based on our empirical results.

\section{CoT Improvement}
\label{sec-method}

\begin{figure*}[tb]
	\centering
	\includegraphics[width=0.98\textwidth]{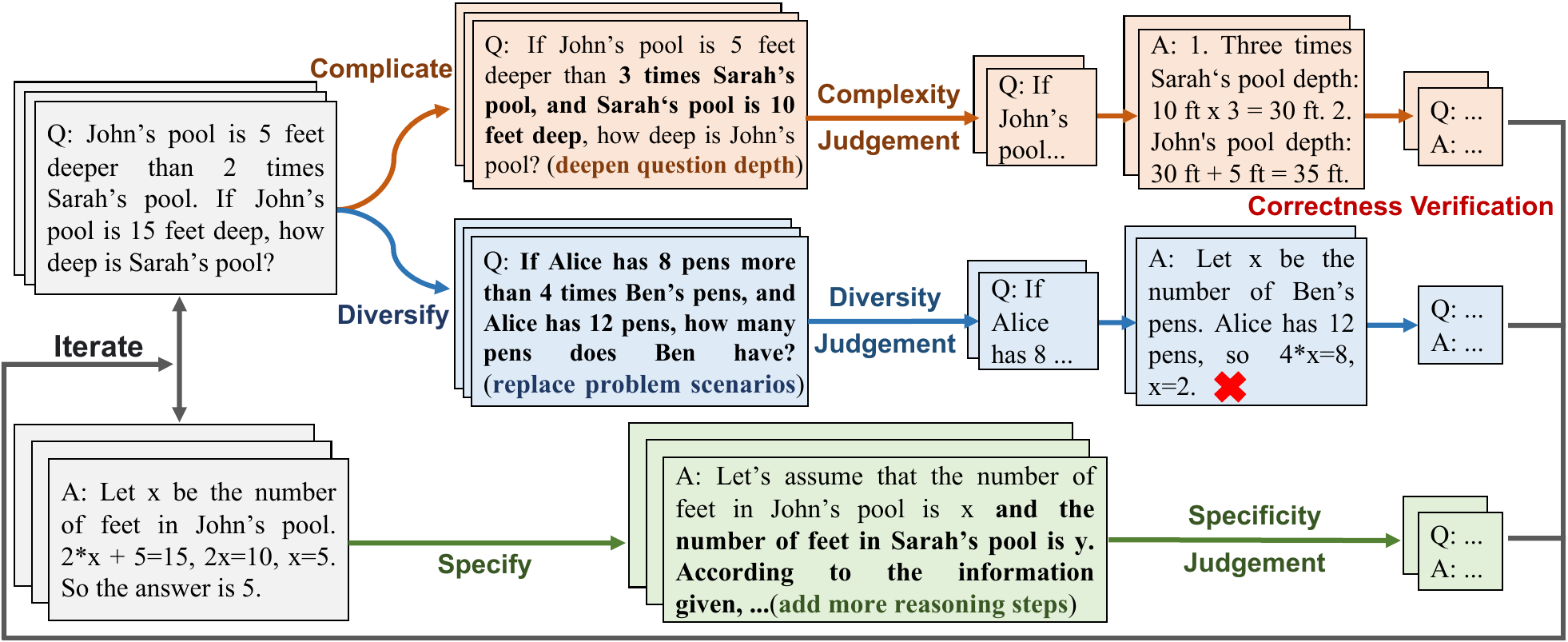}
	\caption{Illustration of our proposed CoTGenius framework for improving CoT prompts.}
	\label{fig-model}
	\vspace{-0.2cm}
\end{figure*}

\subsection{The CoTGenius Framework}
\label{sec-magiccot}

Based on the above empirical studies, we propose a CoT improvement framework, CoTGenius, which introduces three evolution strategies to retrofit existing CoT prompts. Besides, we propose evolutionary success judgement and correctness verification to filter out the failed evolved and erroneous CoT prompts.
The overall illustration of CoTGenius is shown in Figure~\ref{fig-model}.

\subsubsection{CoT Improvement Strategies}

Existing work mostly focuses on improving instructions to enhance the instruction following capability of LLMs~\cite{xu2023wizardlm,Guo-2023-arxiv-Connecting}. Compared to them, CoT improvement requires considering both the question and reasoning steps, as well as the consistency between them. 
According to our empirical studies, we design three CoT improvement strategies using evolutionary methods including complicate, diversify, and specify.
Detailed instructions of three strategies can be found in Table~\ref{tab: evolve approach} in the Appendix~\ref{app: strategies}.

\paratitle{Complicate}. In general, solving complex problems requires a number of reasoning steps, which can better demonstrate the step-by-step reasoning process and is beneficial to stimulate the model's reasoning ability during fine-tuning. Thus, we propose a \emph{complication} evolution strategy, aimed at upgrading the question in CoT prompts into a complex and challenging one involving more reasoning steps. Specifically, we use ChatGPT to complicate the initial question based on complication instructions paired with few-shot in-context demonstrations. In our complication instructions, we adopt two methods to increase the difficulty level of the question:

\textbullet~\textbf{Adding Conditions and Constraints}: Simply adding conditions and constraints to the question can directly lead to an increase of reasoning steps. For example, a math question ``$x^2=16$, what is $x$?'' can be complicated by adding constraints $x>0$.

\textbullet~\textbf{Deepening Question Depth}: In addition to adding constraints, the second method focuses on increasing the depth of the given question, expecting to transform superficial questions into profound ones and further fostering the complexity of CoT. For example, we can deepen the math question ``$x^2=16$, what is $x$?'' into a challenging one ``$x^2=16$, $x>0$, is $x$ a prime number?''.

After complicating the initial questions in CoT prompts, we further use ChatGPT to generate their corresponding CoT reasoning process with more steps compared to the initial reasoning process.

\paratitle{Diversify}. 
Many studies have reported that increasing the diversity of training data can enhance the generalization ability of LLMs~\cite{YuLan-Chat}. Therefore, we propose a \emph{diversity} evolution strategy to expand the question topics of CoT prompts, making our model applicable to more general scenarios. Similarly, we also leverage ChatGPT to execute the diversity evolution by the following two methods:

\textbullet~\textbf{Replacing Problem Scenarios}: 
This method can totally change the background of the original problems with the aim of increasing the topic diversity of the problem formally. For example, a problem scenario ``A pipe takes an hour to fill the tank'' can be changed to ``A car travels from point A to point B at a speed of 60 km/h''.

\textbullet~\textbf{Drawing Inspiration from the Given Question}: 
Besides changing the background, it is more important to diversify the core of a given problem.
For this purpose, ChatGPT is employed to leverage the content of given questions as a source of inspiration to craft completely new questions. For example, the math question ``$x^2=16$, what is $x$?'' can be reformulated into ``The radius of a circle $x$ satisfies $x^2=16$, what is the radius of the circle?''.

After diversifying the questions in the initial CoT prompts, we also utilize ChatGPT to generate their reasoning processes correspondingly.

\paratitle{Specify}. 
As observed in our empirical study about the prompt specificity, CoT prompts with sufficient details can significantly improve the performance of the model.
Therefore, we introduce a \emph{specificity} evolution strategy to insert more details into the CoT reasoning steps while keeping the underlying question unchanged. Specifically, this strategy involves ChatGPT to rewrite the CoT reasoning process corresponding to each question using two methods:

\textbullet~\textbf{Adding More Reasoning Steps}: The first approach incorporates additional reasoning steps into the original CoT prompts, making the reasoning process more detailed and specified. This enhancement aims to provide LLMs with clearer and more comprehensive step-by-step reasoning guidance.

\textbullet~\textbf{Rewriting Existing Reasoning Steps}: The second method will revise existing reasoning steps within CoT prompts to make them more standardized and smooth. In particular, the standardization operation aims to make the CoT prompts consistent with the logic of reasoning-then-answering, while the smoothing operation is to make the reasoning conducted step-by-step rigorously.

In the specificity process, we remain the input question unchanged but improve CoT reasoning steps.
By specifying the CoT rationales, we endeavor to enhance their clarity, details, and effectiveness to guide LLMs in complex reasoning tasks.

\subsubsection{CoT Filtering}

Although we employ ChatGPT to conduct the CoT improvement automatically, it is imperative to ensure that the synthesized CoT prompts align well with the evolution strategies. Besides, the evolving CoT reasoning steps might be inconsistent with the question and contain errors, so we need to verify the correctness of CoT prompts and filter those failed evolved data.
To achieve these goals, we introduce two CoT filtering methods: evolutionary success judgement and correctness verification.

\paratitle{Evolutionary Success Judgement}. In this method, we adopt three powerful LLMs, \ie ChatGPT, Claude, and PaLM, to assess whether the rewritten questions in the complication and diversity evolution, as well as the rewritten CoT reasoning steps in the specificity evolution, successfully achieve the evolution objectives compared to the input data. We only retain those CoT prompts that have undergone successful evolution as our synthesized data via max-voting among the three LLMs.

\paratitle{Correctness Verification}. During the process of complication and diversity evolution, we first upgrade the CoT questions and then utilize ChatGPT to generate reasoning steps for these evolved new questions. Thus, we design correctness verification to assess the consistency between the questions and generated reasoning steps. We also use the three aforementioned LLMs to provide judgments regarding the reasoning correctness, and use the max-voting strategy to retain the correct ones. 

The instructions of evolutionary success judgement and correctness verification can be found in Table~\ref{tab: effect-success-judgement} and Table~\ref{tab: correctness-verification} in the Appendix~\ref{app: filtering}.

\begin{figure}[tb]
	\centering
	\includegraphics[width=0.44\textwidth]{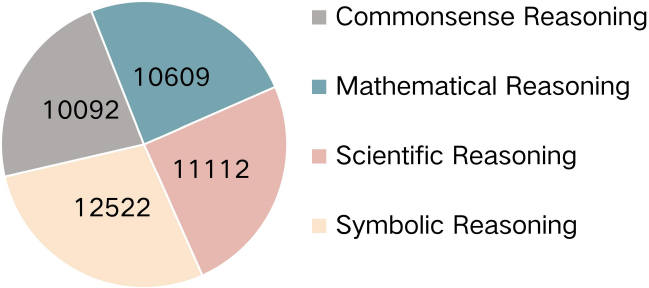}
	\caption{Statistics of synthesized CoT samples.}
	\label{fig-sample}
	\vspace{-0.2cm}
\end{figure}

\subsection{Improved CoT Dataset}
\label{sec-dataset}

We categorize complex reasoning tasks that benefit from the CoT technique into four distinct types, \ie commonsense reasoning, mathematical reasoning, scientific reasoning, and symbolic reasoning. For each task type, we carefully select seed datasets with CoT prompts to serve as the initialization of our improvement process:

\textbullet~\textbf{Commonsense Reasoning}: We adopt StrategyQA~\citeplanguageresource{geva2021did} and Date Understanding~\citeplanguageresource{srivastava2023beyond} as seed datasets.

\textbullet~\textbf{Mathematical Reasoning}: We consider AQUA-RAT~\citeplanguageresource{ling2017program} and GSM8K~\citeplanguageresource{Cobbe-arxiv-2021-Training} as seed datasets.

\textbullet~\textbf{Scientific Reasoning}: We employ ARC-Challenge~\citeplanguageresource{clark2018think}, OpenbookQA~\citeplanguageresource{mihaylov2018can}, and WorldTree~\citeplanguageresource{xie-etal-2020-worldtree} as seed datasets.

\textbullet~\textbf{Symbolic Reasoning}: We include Colored Objects, Tracking Shuffled Objects, and Word Sorting from BIGbench~\citeplanguageresource{srivastava2023beyond} as seed datasets.

We mix up the training sets of the above seed datasets as the input data of our evolution process and perform CoT improvement following the CoTGenius framework. 
These seed datasets are iteratively evolved into new data, with each round of evolution building upon the previous results. 
This iterative process is repeated for four epochs to obtain sufficient data of varying complexity, diversity, and specificity. 
Note that we only keep the newly generated CoT prompts without the seed datasets and randomly shuffle the samples to create the final dataset. In the end, we successfully generate a total of 44,335 samples. The statistics of our dataset are presented in Figure~\ref{fig-sample}.

\begin{figure*}[tb]
	\centering
	\includegraphics[width=1.00\textwidth]{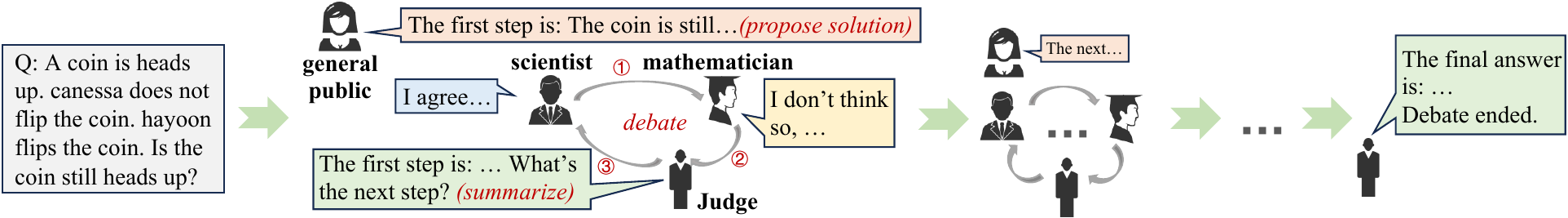}
	\caption{Illustration of our proposed step-level debating strategy.}
	\label{fig-debating}
	\vspace{-0.2cm}
\end{figure*}

\subsection{ChainLM: Fine-tuning LLM on CoT}
\label{sec-magiclm}

To validate the effectiveness of our improved CoT data in enhancing the reasoning capabilities of LLMs, we fine-tune Llama 2-Chat 7B and 13B models~\cite{Touvron-2023-llama2-arxiv} with our dataset for three epochs. We call the fine-tuned model as \emph{ChainLM}. 
Based on our improved CoT data, ChainLM possesses exceptional capabilities in complex reasoning tasks. 
However, we discover that most inaccurate answers usually stem from the errors in intermediate steps and these errors tend to accumulate over steps. 
Thus, to increase the accuracy of intermediate steps, we propose a step-level debating method for improving our ChainLM model.

\paratitle{Step-level Debating}.
Due to the absence of correctness labels for intermediate steps, it is challenging to fine-tune LLMs with process supervision.
Inspired by previous work employing multiple LLMs to discuss for output consensus~\cite{ReConcile,Encouraging-debate,du2023improving}, we propose a step-level debating method to improve the CoT reasoning at each step. Different from prior work focused on discussing about the final solution, our method employs multiple LLMs to engage in debating about each reasoning step for fine-grained consensus.
Specifically, we set one general public and three debaters roles, \ie scientist, mathematician, and judge, which are all built upon our ChainLM model. The instructions for each role are provided in Table~\ref{tab: debater prompt} in the Appendix~\ref{app: debating}. For each question, general public is asked to answer it step by step, while scientist and mathematician are required to debate for each reasoning step. The judge is responsible for summarizing the debate between scientist and mathematician and finally giving an outcome of each reasoning step.
Once a solution consensus is reached at the current reasoning step, general public proceeds to the next step of reasoning, as depicted in Figure~\ref{fig-debating}. To illustrate the entire process, we present an example of step-level debating in Table~\ref{tab: debating example} in Appendix~\ref{app: debating}.

\section{Experiment}
\label{sec-exp}

\begin{table*}
\centering
\resizebox{\textwidth}{!}{
\begin{tabular}{lccccccccc}
\toprule
\multirow{2}{*}{\textbf{Models}} & \multicolumn{2}{c}{\textbf{Commonsense}} & \multicolumn{2}{c}{\textbf{Math}} & \multicolumn{2}{c}{\textbf{Science}} & \multicolumn{2}{c}{\textbf{Symbol}} & \textbf{other domain} \\
\cmidrule(r){2-3}\cmidrule(r){4-5}\cmidrule(r){6-7}\cmidrule(r){8-9}\cmidrule(r){10-10}
 & \textbf{CSQA} & \textbf{SIQA} & \textbf{Math} & \textbf{EleMath} & \textbf{ScienceQA} & \textbf{SciQ} & \textbf{Penguins} & \textbf{O\_Counting} & \textbf{Phrase}\\
\midrule
\textbf{ChatGPT} & 73.55 & 66.84 & 34.10 & 68.52 & 61.92 & 62.80 & 58.62 & 82.50 & 67.50 \\
\textbf{Claude} & - & - & 20.18 & - & - & - & 47.65 & - & -\\
\textbf{Davinci003} & 74.61 & 70.62 & 17.66 & 61.11 & 54.32 & 62.80 & 31.03 & 78.00 & 87.50 \\
\textbf{Davinci002} & 62.98 & 61.87 & 19.10 & 40.48 & 42.81 & 55.70 & 65.52 & 70.00 & 96.25 \\
\midrule
\textbf{LLaMA} & 41.28 & 39.25 & 2.90 & 23.81 & 31.61 & 32.50 & 31.03 & 22.50 & 16.25\\
\textbf{Llama 2-Chat} & 65.35 & 52.19 & 2.50 & 24.88 & 28.87 & 28.70 & 38.93 & 33.50 & 56.25 \\
\textbf{Falcon} & - & - & 2.30 & - & - & - & 24.16 & - & - \\
\textbf{ChatGLM} & - & - & 1.10 & - & - & - & 14.09 & - & -\\
\textbf{Alpaca} & 46.85 & 45.80 & 3.55 & 20.90 & 31.88 & 34.70 & 37.93 & 25.00 & 45.00 \\
\textbf{Vicuna} & 45.95 & 34.49 & 2.60 & 23.54 & 35.66 & 35.10 & 27.59 & 36.00 & 36.25 \\
\textbf{CoT-T5 11B} & \textbf{81.99} & \textbf{69.65} & 1.50 & 26.19 & \underline{43.03} & 44.00 & \underline{44.83} & 32.50 & 77.50 \\
\textbf{WizardLM 7B} & 43.41 & 38.69 & 1.42 & 20.63 & 27.56 & 45.30 & 24.14 & 15.00 & 33.75 \\
\textbf{WizardLM 13B} & 59.87 & 48.93 & \underline{5.30} & \underline{33.33} & 40.83 & \underline{47.00} & 20.69 & 38.00 & 57.50 \\
\midrule
\textbf{ChainLM seed} & 62.33 & 55.17 & 3.80 & 31.22 & 41.01 & 33.40 & 17.24 & 33.50 & 32.50 \\
\textbf{ChainLM 7B} & 65.75 & 54.25 & 3.92 & \underline{33.33} & 42.22 & 42.30 & 41.38 &  \underline{46.00} & \underline{82.50} \\
\textbf{ChainLM 13B} & \underline{68.22} & \underline{55.27} & \textbf{5.38} & \textbf{34.13} & \textbf{43.07} & \textbf{49.40} & \textbf{48.28} &  \textbf{52.50} & \textbf{83.75} \\
\bottomrule
\end{tabular}
}
\caption{Evaluation results at accuracy on nine complex reasoning datasets. \textbf{Bold} and \underline{underline} fonts denote the best and second best methods among open-source models. ``-'' means the model has not been tested on this dataset on their original papers.}
\label{tab: main experiment}
% \vspace{-0.2cm}
\end{table*}

\begin{table*}
\centering
\small
\begin{tabular}{lcccccccc}
\toprule
\multirow{2}{*}{\textbf{Variants}} & \multicolumn{2}{c}{\textbf{Commonsense}} & \multicolumn{2}{c}{\textbf{Math}} & \multicolumn{2}{c}{\textbf{Science}} & \multicolumn{2}{c}{\textbf{Symbol}} \\
\cmidrule(r){2-3}\cmidrule(r){4-5}\cmidrule(r){6-7}\cmidrule(r){8-9}
 &  \textbf{CSQA} & \textbf{SIQA} & \textbf{Math} & \textbf{EleMath} & \textbf{ScienceQA} & \textbf{SciQ} & \textbf{Penguins} & \textbf{O\_Counting}\\
\midrule
\textbf{ChainLM 7B} & 65.75 & 54.25 & 3.92 & 33.33 & 42.22 & 42.30 & 41.38 &  46.00 \\
\midrule
\textbf{w/o CS} & 59.21 & 50.61 & 3.84 & 32.54 & 38.53 & 41.90 & 44.83 &  48.50 \\
\textbf{w/o Math} & 52.83 & 54.91 & 3.30 & 30.69 & 39.34 & 43.50 & 27.59 &  47.00 \\
\textbf{w/o Sci} & 56.67 & 52.87 & 4.42 & 28.84 & 35.75 & 42.30 & 44.83 &  39.00 \\
\textbf{w/o Sym} & 59.79 & 53.68 & 4.42 & 30.16 & 37.77 & 41.40 & 31.03 & 41.50 \\
\bottomrule
\end{tabular}
\caption{Results of ablation study.}
\label{tab: detailed study}
% \vspace{-0.3cm}
\end{table*}

\begin{table*}
\small
\centering
\begin{tabular}{lcccccccc}
\toprule
\multirow{2}{*}{\textbf{Variants}} & \multicolumn{2}{c}{\textbf{Commonsense}} & \multicolumn{2}{c}{\textbf{Math}} & \multicolumn{2}{c}{\textbf{Science}} & \multicolumn{2}{c}{\textbf{Symbol}} \\
\cmidrule(r){2-3}\cmidrule(r){4-5}\cmidrule(r){6-7}\cmidrule(r){8-9}
 & \textbf{CSQA} & \textbf{SIQA} & \textbf{Math} & \textbf{EleMath} & \textbf{ScienceQA} & \textbf{SciQ} & \textbf{Penguins} & \textbf{O\_Counting}\\
\midrule
\textbf{ChainLM 7B} & 65.75 & 54.25 & 3.92 & 33.33 & 42.22 & 42.30 & 41.38 &  46.00 \\
\midrule
\textbf{w/ SC} & 66.09 & 54.55 & 4.59 & 38.88 & 51.17 & 44.30 & 51.72 & 47.00  \\
\textbf{w/ LtM} & 63.72 & 48.36 & 3.54 & 32.80 & 46.58 & 42.50 & 48.28 & 40.50  \\
\textbf{w/ Debating} & 66.50 & 54.60 & 4.76 & 44.44 & 52.47 & 50.50 & 55.17 & 47.50 \\
\bottomrule
\end{tabular}
\caption{A comparison between our step-level debating method and other reasoning methods.}
\label{tab: improvement study}
% \vspace{-0.2cm}
\end{table*}

In this part, we detail the experimental setup and then highlight the main takeaways of our results.

\subsection{Experimental Setup}

\paratitle{Datasets and Metrics}.
As shown in section~\ref{sec-dataset}, we categorize complex reasoning tasks into four types. To further evaluate the performance of our model on these tasks, we select two widely-used datasets for each task type:

\textbullet~\textbf{Commonsense Reasoning}: We select CommonsenseQA~\citeplanguageresource{talmor2019commonsenseqa}, a multiple-choice QA dataset requiring commonsense knowledge to predict the answer, and SocialIQA~\citeplanguageresource{sap2019social}, containing multiple-choice questions for probing emotional and social intelligence in a number of daily situations. We conduct 3-shot evaluation on validation sets of CommonsenseQA (1221 samples) and SocialIQA (1954 samples).

\textbullet~\textbf{Mathematical Reasoning}: We choose MATH ~\citeplanguageresource{hendrycks2021measuring}, a challenging dataset about competition mathematical problems, and Elementary Mathematics from MMLU~\citeplanguageresource{hendrycks2020measuring}, containing elementary mathematical problems. We perform 3-shot evaluation on test set of MATH (5000 samples) since it is pretty challenging for LLMs and zero-shot on test set of Elementary Mathematics (378 samples).

\textbullet~\textbf{Scientific Reasoning}: We choose a multimodal multiple-choice science question dataset ScienceQA~\citeplanguageresource{lu2022learn} and a science exam questions dataset SciQ~\citeplanguageresource{welbl2017crowdsourcing}. These two datasets center around the topics like physics, chemistry, and biology. We evaluate our model in zero-shot manner on test sets of ScienceQA (2224 samples) and SciQ (1000 samples).

\textbullet~\textbf{Symbolic Reasoning}: We select Penguins in a Table and Object Counting from BIGbench~\citeplanguageresource{srivastava2023beyond}. The former is to answer questions about animals in a table, and the latter is to count different types of objects without any choice provided. We also conduct zero-shot evaluation on validation set of Penguins in a Table (29 samples) and 3-shot on Object Counting (200 samples).

Moreover, we also select an out-domain dataset, \ie \textbf{Phrase Relatedness}~\citeplanguageresource{srivastava2023beyond}, which does not fall into the above four categories. This dataset will present models with a phrase ($n$-gram) and ask them to select the most related phrase ($n$-gram) among four choices. Therefore, it can be used to evaluate the semantic reasoning performance of our model. We adopt the training set of Phrase Relatedness (80 samples) for zero-shot evaluation. In our experiments, we compute the accuracy (\%) for all evaluation datasets.

\paratitle{Baselines}.
We compare our ChainLM models with a wide range of existing powerful closed-source and open-source models. Specially, we include four representative closed-source LLMs, including ChatGPT (gpt-3.5-turbo)\footnote{https://openai.com/blog/chatgpt/}, InstructGPT (text-davinci-002/003)~\cite{ouyang2022training}, and Anthropic's Claude model. 
For open-source LLMs, we choose LLaMA (7B)~\cite{Touvron-arxiv-2023-LLaMA}, Llama 2-Chat (7B)~\cite{Touvron-2023-llama2-arxiv}, Falcon (7B)~\cite{falcon40b}, ChatGLM (6B)~\cite{du2022glm,zeng2022glm}, Alpaca (7B)~\cite{alpaca}, Vicuna (7B)~\cite{vicuna2023}, WizardLM (7B and 13B)~\cite{xu2023wizardlm} and CoT-T5 (11B)~\cite{kim2023cot} for comparison. 
It is worth noting that CoT-T5 is a model obtained by fine-tuning T5 on a large-scale CoT collection, which contains 1.88 million CoT rationales across 1060 tasks generated by LLMs. 
To validate the effectiveness of our synthesized CoT data compared to the original seed data, we directly fine-tune Llama 2-Chat (7B) on the mixture of seed datasets as a baseline for evaluation, called ChainLM seed.

\paratitle{Implementation Details}.
We adopt the code of Alpaca~\cite{alpaca} to fine-tune our models. Specifically, we fine-tune Llama 2-Chat on our improved CoT data for three epochs with a learning rate of 2e-5. We set the maximum number of tokens to 4096, and batch size to 128, 256 for Llama 2-Chat 7B and 13B, respectively. For evaluation, we set the temperature to 0.1 for all models to reduce output randomness and set the maximum number of tokens for generation to 512.

\subsection{Results and Analysis}

The performance results of our model and the comparison models are presented in Table~\ref{tab: main experiment}. 
First, we can observe that our ChainLM model substantially outperforms other open-source models. For example, compared to Llama 2-Chat, we achieve significant improvement at accuracy from 24.88 to 33.33 in Elementary Mathematics, 28.87 to 42.22 in ScienceQA, and 33.50 to 46.00 in Object Counting, which demonstrates the effectiveness of fine-tuning on our improved CoT data. Since MATH is composed of exceptionally challenging math problems, open-source models only achieve low performance compared to closed-source models.
Second, CoT-T5 is the most relevant model to us and our model achieves much better results than CoT-T5 except CSQA and SIQA. We hypothesize that commonsense reasoning relies more on world knowledge instead of the reasoning ability of models, while CoT-T5 is trained on extensive NLP tasks to gain knowledge.
It is particularly noteworthy that ChainLM outperforms the baseline fine-tuning on the original seed datasets, which clearly underscores the effectiveness of our CoT improvement framework.
Finally, scaling our fine-tuned model size to 13B obtains a closer or even better results compared to closed-source models on certain datasets. In Phrase Relatedness, our model achieves 83.75 accuracy better than ChatGPT.

\subsection{Ablation Study}

Our ChainLM model is fine-tuned on four types of complex reasoning tasks with CoT prompts, thus it is non-trivial to deeply analyze the impact of each task type on the final model performance. In this part, we conduct ablation study by removing one task type from synthesized CoT data (Section~\ref{sec-dataset}) and then fine-tuning Llama 2-Chat 7B.

Table~\ref{tab: detailed study} presents the results of four model variants. It can be seen that removing any task type generally leads to varying degrees of performance degradation. In most cases, the performance on target task closely correlates with the task type in our fine-tuning data. For example, the model fine-tuned without scientific reasoning data exhibits the lowest accuracy on the ScienceQA dataset among the four variants. It is interesting that we notice an implicit relation between commonsense reasoning and scientific reasoning---removing commonsense reasoning data results in a dramatic accuracy degradation in scientific reasoning, and vice versa.
The reasons might be that some commonsense reasoning problems require scientific knowledge to assist in their solution, and in scientific problems there are numerous commonsense involved in reasoning.
Furthermore, we observe that the performance in symbolic reasoning actually improves after removing some data, which could be attributed to synthesized data interfering with the comprehension of symbols and rules.

\subsection{CoT Reasoning Strategies}

In Section~\ref{sec-magiclm}, we propose a CoT reasoning strategy, step-level debating, employing multiple LLMs to discuss about each reasoning step for consensus.
Therefore, to validate its effectiveness, we compare our method to previous CoT reasoning strategies, \ie self-consistency~\cite{wang2022self} and least-to-most~\cite{zhou2022least}. 
We set 10 paths for self-consistency and three rounds for our step-level debating. 

As we can see from Table~\ref{tab: improvement study}, our step-level debating method outperforms baselines consistently. Although self-consistency also adopts max-voting to achieve consensus among multiple solutions, it only focuses on the final answer without considering intermediate steps. In contrast, our method proceeds to the next reasoning step only when a consensus is reached at the current step. The failure of least-to-most lies in that many problems cannot be accurately decomposed into simpler subproblems. 
Our step-level debating method performs well especially in tasks where errors are prone to occur in intermediate steps, \ie Elementary Mathematics and Penguins in a Table. This significantly illustrates the necessity of considering the consistency during the reasoning process.

\section{Related Work}
\label{sec-rel}

\paratitle{Chain-of-Thought Prompting}.
Chain-of-Thought (CoT)~\cite{wei2022chain,kojima2022large} prompting is an effective solution for solving complex problems by explicitly generating reasoning steps. 
As CoT prompting has a critical effect on improving model performance, many studies~\cite{saparov2022language, wang2023towards} are proposed to further improve CoT prompting technique.
For example, self-consistency~\cite{0002WSLCNCZ23} proposes generating several reasoning paths and then selecting the most consistent answer by voting. Self-verification~\cite{weng2022large} lets LLMs themselves verify their prediction results. Tree-of-Thought~\cite{yao2023tree} is a paradigm that allows LLMs to explore multiple reasoning paths over thoughts by framing the problem as a search over a tree.  
Instead of improving CoT prompting from the perspective of generation methods, our work explores the factors about why CoT prompting works through empirical analysis and designs a data augmentation framework to improve CoT prompting.

\paratitle{Instruction Tuning}. 
Instruction Tuning is an approach to fine-tuning LLMs on a collection of instructions and responses. 
Early work focuses on fine-tuning LLMs on specific NLP tasks. 
T5~\cite{raffel2020exploring} first introduces text-to-text framework to fine-tune models on multiple tasks. Subsequent work such as FLAN~\cite{wei2021finetuned}, FLAN-T5~\cite{chung2022scaling}, and ZeroPrompt~\cite{xu2022zeroprompt} improve LLMs by increasing the number of tasks and carefully designing instructions for the tasks.
To bridge the gap between human queries and synthesized instructions, many studies propose to fine-tune LLMs with open-domain instructions.
Alpaca~\cite{alpaca} is fine-tuned on 52K instruction data generated by Self-Instruct~\cite{wang2022self}. Vicuna~\cite{vicuna2023} collects 70K user-shared ChatGPT conversations from ShareGPT.com for fine-tuning. WizardLM~\cite{xu2023wizardlm} proposes Evol-Instruct to evolve instructions. Our method is most similar to WizardLM, except that we focus on enhancing CoT data and improving the complex reasoning capabilities of LLMs.

\section{Conclusion}
\label{sec-con}

This paper presented a CoT improvement framework, CoTGenius, which encompasses three evolution strategies, \ie complicate, diversify, and specify, alongside two filtering mechanisms: evolutionary success judgement and correctness verification.
We fine-tuned Llama 2-Chat on superior CoT prompts synthesized by our CoTGenius framework, called ChainLM.
Through a series of rigorous experiments, we demonstrate that our model exhibited better performance when confronted with complex reasoning tasks.
To further deal with the accumulative error issue in intermediate reasoning steps, we proposed step-level debating, a collaborative approach where multiple agents engage in discussion for each CoT reasoning step to yield correct answers.
We believe that our data and model can  facilitate future work towards powerful LLMs.

\section{Acknowledgements}
This work was partially supported by National Natural Science Foundation of China under Grant No.62222215, Beijing Natural Science Foundation under Grant No. 4222027 and L233008. Xin Zhao is the corresponding author.
\section{Bibliographical References}\label{sec:reference}

\bibliographystyle{lrec_natbib}
\bibliography{lrec-coling2024-example}

\section{Language Resource References}
\label{lr:ref}

\bibliographystylelanguageresource{lrec_natbib}
\bibliographylanguageresource{languageresource}

\newpage
\appendix
\section*{Appendix}
\label{sec-app}
We provide the implementation details of the CoTGenius framework as supplementary materials.

\section{CoT Improvement Strategies}
\label{app: strategies}

We present the instructions for the three CoT improvement strategies including complicate, diversify, and specify in Table~\ref{tab: evolve approach}.

\begin{table*}[t]
	\small
	\centering
	\begin{tabular}{lp{0.8\textwidth}}
		\toprule
            \multirow{12}{*}{\textbf{Complicate}}&I want you to act as a Question Rewriter. Your objective is to rewrite a given question into a more complex version to make it require more reasoning steps. But the rewritten question must be reasonable, understandable, and answerable by humans. \\
            &\\
            &You SHOULD complicate the given question using the following methods: \\
            &1. Add some more constraints/requirements into \#Given Question\#. \\
            &2. Increase the depth of the \#Given Question\#. \\
            &\\
            &\#Rewritten Question\# must be a solvable problem independent of the \#Given Question\#. \\
            &\\
            &\#Given Question\#:  \\
            &\#Rewritten Question\#: \\
            \midrule
            \multirow{12}{*}{\textbf{Diversify}}&I want you to act as a Question Rewriter. Your objective is to rewrite a given question into a more diverse version. But the rewritten question must be reasonable, understandable, and answerable by humans. \\
            &\\
            &You SHOULD diversify the given question using the following methods: \\
            &1. Replace problem scenarios. \\
            &2. Draw inspiration from the \#Given Question\# to create a brand new question. \\
            &\\
            &\#Rewritten Question\# must be a solvable problem independent of the \#Given Question\#. \\
            &\\
            &\#Given Question\#:  \\
            &\#Rewritten Question\#: \\
            \midrule
            \multirow{13}{*}{\textbf{Specify}}&I want you to act as a Chain-of-Thought Rewriter. Given a question and its Chain-of-Thought answer, your objective is to rewrite the given Chain-of-Thought answer into a more specific version. But the rewritten CoT must be reasonable and have the same answer as the given CoT. \\
            &\\
            &You SHOULD specify the given CoT using the following methods: \\
            &1. Add more reasoning steps to make the reasoning progress more detailed. \\
            &2. Rewrite existing reasoning steps to make them more standardized. \\
            &\\
            &\#Given Question\#:  \\
            &\#Given CoT\#: \\
            &\#Rewritten CoT\#: \\
            \bottomrule
	\end{tabular}
\caption{Instructions of three evolutionary strategies.}
\label{tab: evolve approach}
\end{table*}

\section{CoT Filtering}
\label{app: filtering}
We present the instructions for evolutionary success judgement including complicate, diversify, and specify in Table~\ref{tab: effect-success-judgement}. The instruction for correctness verification is shown in Table~\ref{tab: correctness-verification}.

\begin{table*}[t]
	\small
	\centering
	\begin{tabular}{lp{0.8\textwidth}}
		\toprule
            \multirow{7}{*}{\textbf{Complicate}}&Given two questions, try your best to judge whether \#Question 2\# is more difficult than \#Question 1\#. If \#Question 1\# is more difficult, write 'No'. If \#Question 2\# is more difficult, write 'Yes'.\\
            &\\
            &\#Question 1\#:  \\
            &\#Question 2\#:  \\
            &\#Your Judgement\#: \\
            \midrule
            \multirow{6}{*}{\textbf{Diversify}}&Given two questions, try your best to judge whether \#Question 2\# is different from \#Question 1\#. If the two questions are different, write 'Yes'. Otherwise, write 'No'.\\
            & \\
            &\#Question 1\#:  \\
            &\#Question 2\#:  \\
            &\#Your Judgement\#: \\
            \midrule
            \multirow{8}{*}{\textbf{Specify}}&Given a question and two Chain-of-Thought answers to the question, try your best to judge whether \#CoT 2\# is better than \#CoT 1\#. If \#CoT 2\# is better than \#CoT 1\#, write 'Yes'. If \#CoT 1\# is better than \#CoT 2\#, write 'No'.\\
            & \\
            &\#Question\#:  \\
            &\#CoT 1\#:  \\
            &\#CoT 2\#:  \\
            &\#Your Judgement\#:\\
            \bottomrule
	\end{tabular}
\caption{Instructions of evolutionary success judgement.}
\label{tab: effect-success-judgement}
\end{table*}

\begin{table*}[t]
	\small
	\centering
	\begin{tabular}{p{0.93\textwidth}}
		\toprule
            Given a question and an answer to the question, try your best to judge whether the answer is right or wrong. If it's right, write 'Yes'. If it's wrong, write 'No'. \\
            \\
            \#Question\#:  \\
            \#Answer\#:  \\
            \#Your Judgement\#: \\
            \bottomrule
	\end{tabular}
\caption{Instructions of correctness verification.}
\label{tab: correctness-verification}
\end{table*}

\section{Step-level Debating}
\label{app: debating}

We provide descriptions for the roles of general public, scientist, mathematician, and judge for debater initialization in Table~\ref{tab: debater prompt}. 
To clarify the step-level debating process, we present a step-level debating example of Object Counting in Table~\ref{tab: debating example}.

\begin{table*}[t]
	\small
	\centering
	\begin{tabular}{lp{0.77\textwidth}}
		\toprule
            \multirow{4}{*}{\textbf{general public}} & You are the general public. Given a problem, you should give one step of your Chain-of-Thought answer. For each step, we will have a debating and the judge will decide the final answer for this step. You need to give the next step based on the previous steps until the judge gives the final answer. It must be noted that you can only give one step at a time. \\
            \midrule
            \multirow{3}{*}{\textbf{scientist}} &  You are the scientist. Given a problem and one solving step, you should judge whether the step and discussion are correct. If they are not correct, you should give your reason and your opinion of the correct step.\\
            \midrule
            \multirow{3}{*}{\textbf{mathematician}} & You are the mathematician. Given a problem and one solving step, you should judge whether the step and discussion are correct. If they are not correct, you should give your reason and your opinion of the correct step.\\
            \midrule
            \multirow{4}{*}{\textbf{judge}} & You are the judge. Given a problem and the debating process of one solving step, you should judge which opinion is correct and give the answer of the very step. If you can conclude the final answer directly, repeat the final answer with 'Debate ended.' in the end. If there are choices in the question, give the right choice.\\
            \bottomrule
	\end{tabular}
\caption{Descriptions for the roles of general public, scientist, mathematician, and judge in CoT debating.}
\label{tab: debater prompt}
\end{table*}

\begin{table*}[t]
	\small
	\centering
	\begin{tabular}{lp{0.80\textwidth}}
		\toprule
            \multirow{2}{*}{\textbf{user}} & I have a potato, three heads of broccoli, a cabbage, and two stalks of celery. How many vegetables do I have? \\
            \midrule
            \textbf{general public}  & First, potatoes, broccoli, cabbage, and celery are all vegetables.\\
            \textbf{scientist}  & I agree. \\
            \textbf{mathematician} & I think the first step of reasoning is correct.\\
            \textbf{judge} & Potatoes, broccoli, cabbage, and celery are all vegetables.\\
            \midrule
            \multirow{3}{*}{\textbf{general public}} & I will count the three heads of broccoli as one vegetable. Then, I will count the cabbage as another vegetable. Finally, I will count the two stalks of celery as another vegetable. So, in total, I have three vegetables.\\
            \multirow{4}{*}{\textbf{scientist}} & The count is incorrect. The three heads of broccoli should be counted as separate vegetables, resulting in a total of 4 vegetables. Additionally, the cabbage and celery should also be counted separately, resulting in a total of 6 vegetables. Therefore, the correct count of vegetables is 4 (broccoli heads) + 1 (cabbage) + 2 (celery stalks) = 7 vegetables. \\
            \multirow{4}{*}{\textbf{mathematician}} & I disagree with the count itself. By considering each broccoli head as a separate vegetable, we arrive at 3 vegetables. Additionally, counting the potato, cabbage, and celery separately, we have a total of 7 vegetables. Therefore, the correct count of vegetables is 7, which includes 1 potato, 3 broccoli heads, 1 cabbage, and 2 celery stalks.\\
            \multirow{6}{*}{\textbf{judge}} & The last reasoning step is correct in recognizing the need to count the number of vegetables. However, the counts of general public and scientist are incorrect. By considering each broccoli head as a separate vegetable, we arrive at 3 vegetables. Additionally, counting the potato, cabbage, and celery separately, we have a total of 7 vegetables. Therefore, the correct count of vegetables is 7, which includes 1 potato, 3 broccoli heads, 1 cabbage, and 2 celery stalks. Debate ended.\\
            \bottomrule
	\end{tabular}
\caption{An example of step-level CoT debating.}
\label{tab: debating example}
\end{table*}

\end{document}